\documentclass[10pt, a4paper]{article}
\usepackage{lrec2016}

\usepackage{multibib}
\newcites{languageresource}{Language Resources}
\usepackage{graphicx}

\usepackage{epstopdf}
\usepackage[latin1]{inputenc}
\usepackage{amsmath}
\usepackage{booktabs} 
\usepackage{multirow}
\usepackage{xcolor}
\usepackage{url}
\title{Analysis of Socially Unacceptable
Discourse with Zero-shot Learning}
\name{Mohamed Rayane GHILENE, Dimitra NIAOURI, Michele LINARDI, Julien LONGHI}

\address{ENSEA Engineering School, ETIS UMR-8051 CY Cergy Paris Universit{\'e}, AGORA CY Cergy Paris Universit{\'e} \\
         rayane.ghilene@ensea.fr, \{michele.linardi, dimitra.niaouri, julien.longhi\} @cyu.fr\\}
        
\abstract{
Socially Unacceptable Discourse (SUD) analysis is crucial for maintaining online positive environments. 
We investigate the effectiveness of Entailment-based zero-shot text classification (unsupervised method) for SUD detection and characterization by leveraging pre-trained transformer models and prompting techniques. 
The results demonstrate good generalization capabilities of these models to unseen data and highlight the promising nature of this approach for generating labeled datasets for the analysis and characterization of extremist narratives.
The findings of this research contribute to the development of robust tools for studying SUD and promoting responsible communication online. \textit{Accepted for publication in the International Conference on CMC and Social Media Corpora for the Humanities (University C\^ote d'Azur, France, 2024).} \\
\newline 
 }

\begin{document}

\maketitleabstract

\section{Introduction}

Large Language Models (LLMs) have showcased remarkable capabilities in Natural Language Processing (NLP) thanks to their contextual understanding of word embeddings, which have proven to be useful in multiple tasks, including text answering, text generation, and data annotation. 
LLMs have also shown potential for text classification tasks such as sentiment analysis \cite{zhang2022adaptive} by leveraging prompt learning.

In recent years, the spread of Socially Unacceptable Discourse (SUD), including hate speech and toxic comments, in various online platforms has underscored the need for novel tools able to identify and characterize these harmful discourses. 
However, developing robust automatic SUD  classifiers comes with multiple challenges. 
For instance, the challenge of adopting a universal definition of SUD due to the numerous discourse characterizations causes ambiguity and subjectivity in corpora adopted to train Machine Learning (ML) models~\cite{DBLP:journals/ipm/KoconFGPKK21}.
Such a scenario poses significant challenges to the creation of well-annotated SUD text corpora that can extensively evaluate the quality of state-of-the-art classification models in large-scale scenarios.

{\noindent \bf SUD Classification challenges} LLMs have obtained state-of-the-art performance in SUD text classification tasks.
In this sense, Carneiro et al. (2023) have recently shown that Masked Language Models (MLM) represent a strong candidate classifier option in multiple online annotated corpora.
At the same time, Causal Language Models (CLM), which are LLM variants specifically trained to learn cause-effect dynamics (usually adopted by generative AI) can also be successfully leveraged in hate speech classification~\cite{DBLP:conf/emnlp/ZhangCY23}.

Despite the effectiveness of these models, we note that LLMs lack generalizability in SUD modeling due to their nature, which consists of understanding statistical relationships between words rather than modeling the meaning of these words within their context.
Zhang et al. (2023)  show that LLMs often obtain solid classification performance in the presence of language stereotypes (e.g., race or religion-related).

On the other hand, in a large-scale context~\cite{carneiro2023studying}, where heterogeneous subdomains of toxic speech require to be differentiated (i.e., multi-class classification) LLMs are not capable of providing accurate classification due to the presence of overlapping characteristics among different speech classes, but also for the presence of subtle linguistic nuances that require to understand the underlying context to be detected.

Moreover, the annotation schema plays a crucial role in the supervised model training. 
Often, SUD annotation is subjective and prone to biases resulting from the annotator's background,  gender, first language, age, and education~\cite{al-kuwatly-etal-2020-identifying}. For instance, significant disagreement among annotators from different cultures regarding the offensiveness of online language has been reported in previous studies ~\cite{thorn-jakobsen-etal-2022-sensitivity}.

{\noindent \bf Contribution} In this work, we present a novel SUD analysis framework, in which we adopt a zero-shot learning paradigm for the automatic detection and characterization of SUD in a large-scale context composed of multiple heterogeneous corpora. 
Specifically, we leverage natural language inference (NLI) pre-trained models to perform SUD inference (a.k.a. entailment) in text instances.
The benefit of this approach is two-fold: first, we do not require data complying with a fixed annotation schema, which may be prone to human bias, second, it will permit to leverage human
expertise for hypothesis engineering and validation~\cite{goldzycher2022hypothesis}, where the users can incorporate their understanding of a specific domain or field to guide the classification process.

\section{SUD Framework based on Natural Language Inference}

In our solution, we leverage Natural language Inference (NLI) pre-trained models, which are a specific type of NLP models trained to understand the relationship between two pieces of text, namely the \textit{premise} and the \textit{hypothesis} (a new text, potentially related to the premise).

\subsection{Entailment Template}

To define premise-hypothesis entailment, we follow a methodology similar to the one proposed for text classification~\cite{gera2022zeroshot}, adapted to perform unsupervised data labeling.

\begin{figure}[h]
  \centering
  \includegraphics[width=0.35\textwidth]{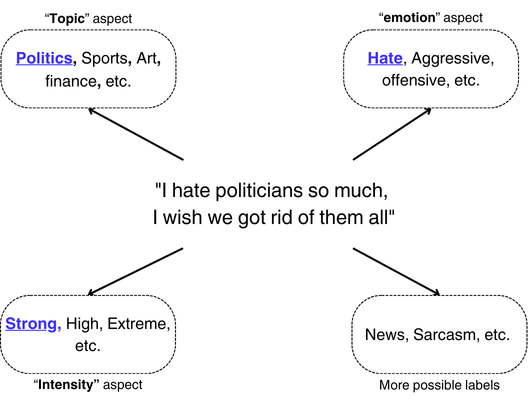}
  \caption{A piece of text can be assigned labels that describe the different aspects of the text. Relevant labels are in blue. Different characterizations of a hateful stance are at the basis of hate speech analysis~\cite{DBLP:conf/emnlp/QianBLBW19}.}
  \label{fig:ZSL}
\end{figure}

In this regard,  we showcase an illustrative example, drawing inspiration from prior research ~\cite{DBLP:conf/emnlp/YinHR19} which we have tailored to SUD analysis, as depicted in Figure~\ref{fig:ZSL}. Here, a hateful premise can be assigned to different labels (hypothesis) according to the perspective under the lens (sentiment, tone of the speech, topics, etc.).

We thus propose an entailment-based framework, where we couple each premise (text item in a corpus) with a hypothesis stating which class it belongs to. 
We construct a pair (text item/hypothesis) for each possible SUD class present in the annotation schema of the dataset.
Constructed pairs become the input of an NLI model that infers a confidence (entailment) score.
In this respect, we consider the output of the softmax layer~\footnote{In our case, the softmax layer takes a textual feature vector (learned by the model)  of real-valued numbers, transforming it into a probability distribution over a set of possible categories (hypothesis).} of an NLI model, where for each hypothesis a probability is assigned between $0$ (contradictory hypothesis) and $1.0$ (entailed hypothesis).

In Table~\ref{tab:entailment} we report an entailment example that we obtain using a zero-shot learning paradigm to perform an unsupervised premise/hypothesis entailment.
Note that a hypothesis is composed of a prefix and a candidate label arbitrarily chosen by the user. 

\begin{table}[tb]
    \centering
    \resizebox{\columnwidth}{!}{%
    \begin{tabular}{|p{4.5cm}|c|c|c|}
        \hline
        \multirow{2}{*}{Premise (t)} & \multirow{2}{*}{Hypotheses} & Candidate  &  Entailment  \\
        &  & Labels & Score \\
        \hline
        \multirow{3}{4.5cm}{what's the difference between a pencil arguing and a woman arguing a pencil has a point} & This example is   & hate & \textbf{0.43} \\ \cline{2-4}
        &  This example is   & offensive & 0.35 \\ \cline{2-4}
        &  This example is   & toxic & 0.22 \\
        \hline
    \end{tabular}%
    }

    \caption{Entailment-based zero-shot classification. For every text t (premise) in the dataset, we create multiple hypothesis by considering several known SUD labels.}
    \label{tab:entailment}
\end{table}


\subsection{Entailment Models}

To perform zero-shot entailment-based text classification on the SUD data, we use models trained specifically for natural language interference (NLI). 
Such models are pre-trained on the MNLI (Multi-Genre Natural Language Inference) dataset~\cite{williams2018broadcoverage} which is a large collection of sentence pairs used to evaluate models on their ability to understand entailment between sentences. 

It contains over 433,000 sentence pairs in English, drawn from ten different genres of written and spoken text, including news articles, fiction, and conversations. Each pair consists of a premise sentence (source) and a hypothesis sentence (target). 

Models trained on the MNLI dataset have the ability to generalize well to different types of textual data, thanks to the diversity of genres they have encountered in the training procedure. For the SUD classification task, we use the following models: 

\begin{itemize}

\item \textbf{Roberta-large-mnli}, BERT~\cite{devlin2019bert}, which is a transformer-based language model pre-trained on English text using a masked language modeling (MLM) objective and fine-tuned on the Multi-Genre Natural Language Inference (MNLI) corpus.

\item \textbf{Bart-large-mnli}~\cite{lewis2019bart}, which is a transformer encoder-decoder (seq2seq) model with a bidirectional (BERT-like) encoder and an autoregressive (GPT-like) decoder. BART is pre-trained by corrupting text with an arbitrary noising function and learning a model to reconstruct the original text.
In our work, we consider the model "facebook/bart-large-mnli", BART version pre-trained on MNLI \cite{williams2018broadcoverage} dataset, for Entailment-based Zero shot classification.

We also consider models trained on other NLI datasets:

\item \textbf{xlm-roberta-large-xnli-anli}, is a variant of the XLM-RoBERTa architecture proposed in \cite{conneau2020unsupervised}, fine-tuned on the XNLI (Cross-lingual Natural Language Inference) \cite{conneau2018xnli} and ANLI (Adversarial Natural Language Inference) \cite{williams2020anlizing} datasets. Its primary application is in cross-lingual natural language inference, which involves determining the relationship (such as entailment, contradiction, or neutrality) between pairs of sentences across multiple languages.

\item \textbf{MoritzLaurer/mDeBERTa-v3-base-xnli-multilingual-nli-2mil7}, multilingual natural language inference (NLI) model based on the mDeBERTa-v3 architecture, fine-tuned on a combination of the XNLI dataset and an additional multilingual NLI dataset with 2.7 million examples. The mDeBERTa-v3 architecture enhances its performance by incorporating improvements in transformer design, such as disentangled attention and enhanced mask decoder.

\end{itemize}

\section{Empirical Evaluation}

\begin{table*}[tbp]
\centering

\resizebox{15cm}{!}{%
\begin{tabular}{|c|c|c|c|c|}
\hline
\textbf{Dataset} & \textbf{Source} & \textbf{Sample type} & \textbf{\# Samples} & \textbf{Labels} \\ \hline
Davidson & ~\cite{davidson2017automated} & Tweets & 25,000 & hate, offensive, neither   \\ \hline
Founta & ~\cite{founta2018large} & Tweets & 100,000 &  abusive, hate, neither \\ \hline
Fox & ~\cite{gao2018detecting} & Threads & 1,528 & hate, neither \\ \hline
Gab & ~\cite{DBLP:conf/emnlp/QianBLBW19} & Posts & 34,000 & hate, neither \\ \hline
Grimminger & ~\cite{Grimminger} & Tweets & 3,000 & hate, neither \\ \hline
HASOC2019 & ~\cite{mandl2019overview} & Facebook, Twitter posts & 12,000 & hate, offensive, profane, neither \\ \hline
HASOC2020 & \cite{mandl2020overview} & Facebook posts & 12,000 & hate, offensive, profane, neither \\ \hline
Hateval & ~\cite{basile2019semeval} & Tweets & 13,000 & hate, neither \\ \hline
Olid & ~\cite{zampieri-etal-2019-predicting} & Tweets & 14,000 & offensive, neither \\ \hline
Reddit & ~\cite{yuan2022detect} & Posts & 22,000 & hate, neither \\ \hline
Stormfront & ~\cite{de2018hate} & Threads & 10,500 & hate, neither \\ \hline
Trac & ~\cite{kumar2018aggression} & Facebook posts & 15,000 & aggressive, neither \\ \hline
\end{tabular}%
}
\caption{Summary of datasets~\cite{carneiro2023studying}}
\label{tab:datasets}
\end{table*}

To validate our solution, we perform zero-shot entailment-based classification on several publicly available datasets ~\cite{carneiro2023studying}. Below, we introduce the datasets employed and the results acquired.
For the sake of reproducibility, the implemented source code used in the evaluation is publicly available on a public repository~\footnote{\url{https://github.com/rayaneghilene/ARENAS_Automatic_Extremist_Analysis/tree/main/Entailment_framework}}.

\subsection{Datasets}
We conducted our evaluation in $12$ publicly available datasets containing up to $12$ different classes of SUD~\cite{carneiro2023studying}. 
In Table~\ref{tab:datasets} we report a detailed overview of the English datasets considered in our study.

\subsection{Evaluation of SUD Classifiers}


The first goal of our evaluation is to compare the entailment models (unsupervised) with the results we obtain adopting a supervised classifier that has been specifically trained over the annotation schema provided in each dataset.

Such experiment will permit us to answer the question:  \textbf{\textit{How performance of an elastic and unsupervised method that does not rely on prior SUD knowledge (i.e. the entailment-based zero-shot learning) compare to the ones of a classifier trained over SUD knowledge? }}
For this latter, we consider a state-of-the-art MLM, namely BERT (Bidirectional Encoder Representations from Transformers)~\cite{devlin2019bert} 

Note that Masked language models (MLMs), like BERT, are DL models trained to reconstruct masked tokens within the input sequence. 
Specifically, MLMs learn to predict the original vocabulary identity of a masked word, relying solely on its contextual cues. The significant advantage of those models lies in their bidirectional context, considering both preceding and subsequent tokens during the prediction process.  
In this work, we fine-tune BERT with the available SUD labels in each evaluated dataset.

We also consider a \textit{shallow} learning baseline such as Logistic Regression (LR)~\cite{grimm1995reading} applied to the numerical representation of tokenized text (text vectorization)~\footnote{\url{https://keras.io/api/layers/preprocessing_layers/text/text_vectorization/}}. 

Note that the entailment models we adopt were not pre-trained by leveraging any available ground truth of SUD, and thus they are unsupervised methods in that respect. 
We base our comparison on the macro F1 score, which is an averaging method for the F1 score that's recommended when working with class imbalance. F1 score is a harmonic mean that combines two performance measures for text classifiers: precision (P) and recall (R). These metrics are computed as follows: 
\(\textbf{R} = \frac{\text{True Positives}}{\text{True Positives} + \text{False Negatives}}\) and \(\textbf{P} = \frac{\text{True Positives}}{\text{True Positives} + \text{False Positives}}.\) \\  

The F1 score is calculated based on these metrics as  
\(\textbf{F1 Score} = 2 \cdot \frac{\text{P} \cdot \text{R}} {\text{P} + \text{R}}\). And then the macro F1 score is computed as \(\textbf{Macro F1} = \frac{1}{C} \cdot \sum_{i=1}^{C} (F1_i)\), where C is the total number of classes. 

\subsubsection{Template Selection}

\begin{table}[tb]
\centering
\resizebox{8.5cm}{!}{
\begin{tabular}{|l|c|c|c|c|}
\hline
\textbf{Hypothesis Testing} & \textbf{roBERTa} & \textbf{BART} & \textbf{mDeBERTa} & \textbf{XLM-roBERTa} \\
\hline
this text contains \{\} speech.         &	\textbf{45.7}  & 27.6 & 30.5 & \textbf{40.9} \\ \hline
this text conveys \{\} speech.          &	40.8           & 34.7 & 29.6 & 35.8 \\ \hline
this text reflects \{\} speech.         &	38.3           & 35.5 & 35.3 & 33.8 \\ \hline
this text shows \{\} speech.            &	35.1           & 38.5 & 27.6 & 35.7 \\ \hline
this text implies \{\} speech.          &	33.2           & 39.6 & 29.1 & 32.1 \\ \hline
this text reveals \{\} speech.          &	37.8           & 41.6 & 28.1 & 32.8 \\ \hline
this text exhibits \{\} speech.         & 	38.8           & 33.3 & 24.2 & 40.4 \\ \hline
this text portrays \{\} speech.         &	33             & 36.3 & 34.6 & 31.6 \\ \hline
this text discusses \{\} speech.        & 	34.8           & 37.9 & 38.9 & 34.5 \\ \hline
this text addresses \{\} speech.        &	34.2           & 38   & 38.3 & 37.1 \\ \hline
this text illustrates \{\} speech.      &  	35.9           & 43   & 34.2 & 32.2 \\ \hline
this text expresses \{\} speech.        &	44.5           & 35.7 & 37.3 & 32.9 \\ \hline
this text articulates \{\} speech.      &	45.1           & 42.5 & 35.8 & 31   \\ \hline 
this text suggests \{\} speech.         &	30.1           & 38.6 & 31.6 & 32.8 \\ \hline
this text narrates \{\} speech.         & 	43.2           & 40.5 & \textbf{38.4} & 35.1 \\ \hline
this text questions \{\} speech.        &	32.6           & 42   & 16.4 & 28.6 \\ \hline
this text demonstrates \{\} speech.     & 	35             & 42.2 & 24.7 & 31.5 \\ \hline
this text supports \{\} speech.         & 	22.6           & \textbf{44,4} & 30.3 & 31.9 \\ \hline
this text has \{\} speech.              & 	41.1           & 32.5 & 12.9 & 39.3 \\ 

\hline
\end{tabular}
}
\caption{Hypothesis Testing F1 Scores}
\label{table:Hypothesis-table}
\end{table}


\begin{table*}[tb]
\centering

\resizebox{15cm}{!}{
\begin{tabular}{|l|c|c|c|c|c|c|}
    \hline
      & \multicolumn{2}{|c|} {\textbf{Supervised SUD classification}} & \multicolumn{4}{|c|}{\textbf{Unsupervised SUD classification (entailment-based})}\\ 
     \hline
     \textbf{Dataset}& BERT &  {LR} & Bart-large-mnli & Roberta-large-mnli &xlm-roBERTa& mDeBERTa  \\
     \hline
    Davidson   & 73   & {69.5} & {47.3}         & {44.7}        &41.5 &39.9  \\     \hline
    Founta     & 70.1 & {73.7} & {57.4}         & {57.5}        &42.8 &36.1  \\     \hline
    Fox        & 47.8 & {69.7} & {56.1}         & {55.2}        &52.5 &48.7  \\     \hline
    Gab        & 87.5 & {89.0} & {64.7}         & {67.1}        &58.3 &55.4   \\     \hline
    Grimminger & 51.9 & {50.4} & \textbf{52.5}  & \textbf{56.1} &48.8 &38.5   \\     \hline
    HASOC2019  & 32.9 & {39.9} & {27.5}         & {30.9}        &17.8 &25.8  \\     \hline
    HASOC2020  & 41.7 & {52.5} & {36.7}         & \textbf{42.7} &20.4 &26.4  \\     \hline
    Hateval    & 63.6 & {70.6} & {59.7}         & {61.4}        &57.2 &54.6  \\     \hline
    Olid       & 65.6 & {71.9} & {61.6}         & {61.5}        &52.1 &55.5  \\     \hline
    Reddit     & 81.7 & {83.0} & {56.3}         & {58}          &50.9 &46     \\     \hline
    Stormfront & 66.9 & {68.4} & {62  }         & {62.6}        &55.2 &51.3        \\     \hline
    Trac       & 67.1 & {69.2} & {52.1}         & {64.2}        &61.7 &55.5  \\     \hline
    \end{tabular}
    }
    \caption{Macro F1 Score (\%) of supervised SUD classification VS  Entailment-based unsupervised SUD classification with the NLI models.}
    \label{Entailment-results}
\end{table*}

In our evaluation, we note that hypothesis construction plays a crucial role in NLI model performance, which has a sensitive and different impact on the considered NLI models, each adopting a different Token masking procedure at the pre-training stage. 

In Table~\ref{table:Hypothesis-table}, we report the hypothesis templates we consider in our work.
In detail, we have tested different active parts, i.e., the verb in the formulation, noticing a remarkable impact (+/- 20 in average F1 score) on average SUD classification performance that we report for each model in Table~\ref{table:Hypothesis-table}.
We observe that the four considered NLI models reach the best F1 score using three different hypothesis templates, which we use in the remaining part of the evaluation.

In the same manner, we note that using the word \textit{neither} in the hypothesis template does not provide any contextual information to the inference phase of the neutral class, resulting in sensibly low classification performance.
We obtain the best performance using the term \textit{neutral speech} in the hypothesis instead of the word \textit{neither} found in each dataset annotation schema (see Table~\ref{tab:datasets}).

\subsubsection{Results and Discussion} 

We report experimental results in Table~\ref{Entailment-results}. As expected, entailment-based model classification shows slightly lower performance when using entailment models compared to a pre-trained MLM.
However, this is not the case for all the datasets, in the Grimminger dataset, our approach outperforms the supervised counterparts, showing a better ability in considering the discourse context at the entailment stage, rather than leveraging correlations among text items in the training set, as in the case of the supervised counterparts.

Furthermore, Roberta-large-mnli and Bart-large-mnli exhibit overall better performance than xlm-roBERTa and mDeBERTa, suggesting that pre-training over the MNLI dataset, which covers a wide range of different spoken and written text is a more suitable choice for SUD analysis.

It is also important to note that such results are similar to the ones obtained by \cite{gera2022zeroshot} when performing zero-shot entailment on other types of text classification.
To the best of our knowledge, we are the first to adopt such techniques in SUD analysis.

To conclude, we also observe that there is no clear winner among the supervised classifiers, and a simple Logistic Regression represents an effective solution in the majority of the datasets.

\begin{table}[tb]
\centering

\resizebox{8.5cm}{!}{
\begin{tabular}{|l|p{1.5cm}|p{1.5cm}|p{1.5cm}|p{1.5cm}|}
    \hline
     \textbf{Dataset} & Bart-large-mnli & Bart-large-mnli + Mask & RoBERTa-large-mnli & RoBERTa-large-mnli + Mask  \\
     \hline
    Davidson   & 47.3 & 40.3 & 44.7 & 42.5  \\      \hline
    Founta     & 57.4 & 53   & 57.5 & 49.8  \\      \hline
    Fox        & 56.1 & 55.5 & 55.2 & 57    \\      \hline
    Gab        & 64.7 & 61.4 & 67.1 & 66.6  \\      \hline
    Grimminger & 52.5 & 50.5 & 56.1 & 56.4  \\      \hline
    HASOC2019  & 27.5 & 23.3 & 30.9 & 29.8  \\      \hline
    HASOC2020  & 36.7 & 28.6 & 42.7 & 37.3  \\      \hline
    Hateval    & 60.8 & 58.6 & 61.4 & 61.3  \\      \hline
    Olid       & 61.6 & 59.5 & 61.5 & 61.8  \\      \hline
    Reddit     & 56.3 & 53.6 & 58   & 59.8  \\      \hline
    Stormfront & 62   & 59.1 & 62.6 & 62.6  \\      \hline
    Trac       & 52.1 & 47.6 & 64.2 & 63.4  \\      \hline  
    \end{tabular}
    }
    \caption{\textbf{Zero-shot text classification with token masking} For each zero-shot entailment model and dataset, we compare the macro F1 score of the off-the-shelf model to its score when performing token masking. 
    }
    \label{tab:token-masking}
\end{table}

{ \noindent \bf Mitigating biases in the classification} To further reduce user bias that may occur in the definition of the hypothesis we adopt GloVe~\cite{pennington2014glove} token masking. 

This procedure consists of masking tokens highly correlated with the label used in the hypothesis, causing the models to rely on the context provided by the remaining part of the speech in the classification task.

For each text, we mask the tokens with the highest GloVe similarity to the class name following the idea proposed in~\cite{gera2022zeroshot}. 

For example, when classifying \textit{offensive} SUD, the words correlated to offensive language will be masked in the text.

The experimental results reported in Table~\ref{tab:token-masking} show that the effect of token masking comes only with a slight performance decrease (in most datasets) compared to the results obtained by the entailment models off-the-shelf. 
Such results suggest how entailment-based SUD classification can not only leverage class stereotypes, but it can potentially leverage the remaining part of the speech.

\section{Conclusion and Future Work}

This paper investigates the effectiveness of zero-shot entailment using NLI models for SUD classification. 

Through preliminary experimentation, these models showcased generalization capabilities comparable with supervised counterparts. 
Such a scenario highlights the entailment-based model's potentiality to exploit contextual information in the text rather than learning intra-class correlation using a fixed annotation schema, which may be sensitive to stereotypes of certain kinds of SUD. 

The preliminary results we obtained motivate several future work directions.
First, we would like to explore how to effectively learn templates that allow linguists to use semantically richer and unstructured annotation schemes, also studying scalability issues and tradeoffs of large entailment hypothesis spaces.
We believe that such capability can support supervised learning models currently adopted in SUD analysis to reduce the impact of annotator bias and sensitivity to class stereotypes.

This result will be a valuable advance for the CMC corpora community and work in corpus linguistics, allowing synergies between AI and corpus linguistics researchers.

\section{References}

\footnotesize
\bibliographystyle{lrec2016}
\bibliography{bibliography.bib}

\end{document}